\documentclass[final]{cvpr}

\usepackage{times}
\usepackage{epsfig}
\usepackage{graphicx}
\usepackage{amsmath}
\usepackage{amssymb}
\usepackage{float}
\usepackage{multirow}
\usepackage{enumitem}
\usepackage{graphicx}
\usepackage{booktabs}
\usepackage{multirow}

\usepackage[pagebackref=false,breaklinks=true,letterpaper=true,colorlinks,urlcolor=blue,citecolor=blue,linkcolor=blue,bookmarks=false]{hyperref}

\begin{document}
	
	\title{Parser-Free Virtual Try-on via Distilling Appearance Flows}
	
	\author{Yuying Ge$^1$ \quad
		Yibing Song$^{2\ast}$ \quad 
		Ruimao Zhang$^3$ \quad
		Chongjian Ge$^1$ \quad 
		Wei Liu$^{4}$ \quad 
		Ping Luo$^1$\\
		{$^1$The University of Hong Kong} \quad \quad {$^2$Tencent AI Lab} \\
		{$^3$The Chinese University of Hong Kong (Shenzhen)} \quad {$^4$Tencent Data Platform} \\
		\tt\small{\{yuyingge, rhettgee\}@hku.hk \quad pluo@cs.hku.hk \quad yibingsong.cv@gmail.com}\\
		\tt\small{ruimao.zhang@ieee.org \quad wl2223@columbia.edu}
	}
	
	\renewcommand{\thefootnote}{\fnsymbol{footnote}}
	\footnotetext[1]{Y. Song is the corresponding author. This work is done when Y. Ge is an intern in Tencent AI Lab. The code is available at \url{https://github.com/geyuying/PF-AFN}.}

	\begin{figure}[htb]
		\twocolumn[{
			\renewcommand\twocolumn[1][]{#1}%
			\maketitle
			\vspace{-25pt}
			\begin{center}
				\centering
				\includegraphics[width=1.0\textwidth]{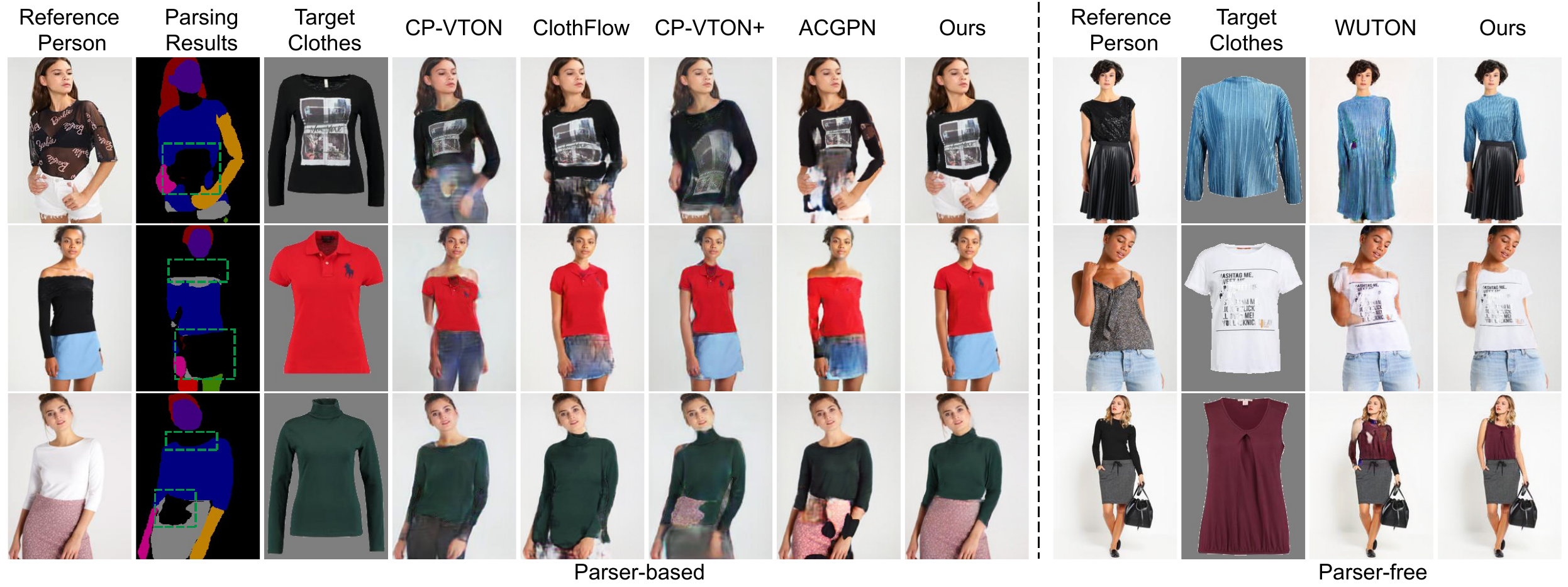}
				\vspace{-8mm}
			\end{center}
			\caption{Comparing our method with the recent state-of-the-art parser-based try-on methods (left) and an emerging parser-free method (right). On the left, we highlight the inaccurate segmentation regions in green boxes, which mislead existing parser-based methods such as CP-VTON \cite{cpvton}, ClothFlow \cite{clothflow}, CP-VTON+ \cite{cpvton_plus}, and ACGPN \cite{ACGPN}
				to produce wrong results. On the right, the first parser-free method WUTON \cite{parser_free} was proposed recently, but its image quality is bounded by the fake images produced by the parser-based method, because \cite{parser_free} simply trained a ``student'' network to mimic the parser-based method using knowledge distillation.
				We see that our approach achieves significantly better image quality than previous state-of-the-art methods, without relying on human segmentation. 
			}
			\label{fig:compare}
			\vspace{20pt}
		}
		]
	\end{figure}
	
	
	\begin{abstract}
		Image virtual try-on aims to fit a garment image (target clothes) to a person image.
		Prior methods are heavily based on human parsing. However, slightly-wrong segmentation results would lead to unrealistic try-on images with large artifacts.
		%
		%
		A recent pioneering work employed knowledge distillation to reduce the dependency of human parsing, where the try-on images  produced by a parser-based method are used as supervisions to train a ``student'' network without relying on segmentation, making the student mimic the try-on ability of the parser-based model. However, the image quality of the student is bounded by the parser-based model.
		To address this problem, we propose a novel approach,  ``teacher-tutor-student'' knowledge distillation, which is able
		to produce highly photo-realistic  images without human parsing, possessing several appealing advantages compared to prior arts.
		(1) Unlike existing work, our approach treats the fake images produced by the parser-based method as ``tutor knowledge'',  where the artifacts can be corrected by real ``teacher knowledge'', which is extracted from the real person images in a self-supervised way.
		(2) Other than using real images as supervisions, we formulate knowledge distillation in the try-on problem as distilling the appearance flows between the person image and the garment image, enabling us to find accurate dense correspondences between them to produce high-quality results.
		(3) Extensive evaluations show large superiority of our method (see Fig.~\ref{fig:compare}).
		
	\end{abstract}

	\section{Introduction}
	Virtual try-on of fashion image is to fit an image of a clothing item (garment) onto an image of human body. This task has attracted a lot of attention in recent years because of its wide applications in e-commerce and fashion image editing.
	Most of the state-of-the-art methods such as VTON \cite{viton}, CP-VTON \cite{cpvton}, VTNFP \cite{vtnfp}, ClothFlow \cite{clothflow},  ACGPN \cite{ACGPN}, and CP-VTON+ \cite{cpvton_plus} were relied on human segmentation of different body parts such as upper body, lower body, arms, face, and hairs, in order to enable the learning procedure of virtual try-on.
	%
	However, high-quality human parsing is typically required to train the try-on models, because slightly wrong segmentation would lead to highly-unrealistic try-on images, as shown in Fig.\ref{fig:compare}.
	
\begin{figure}[t]
		\begin{center}
			\includegraphics[width=\linewidth]{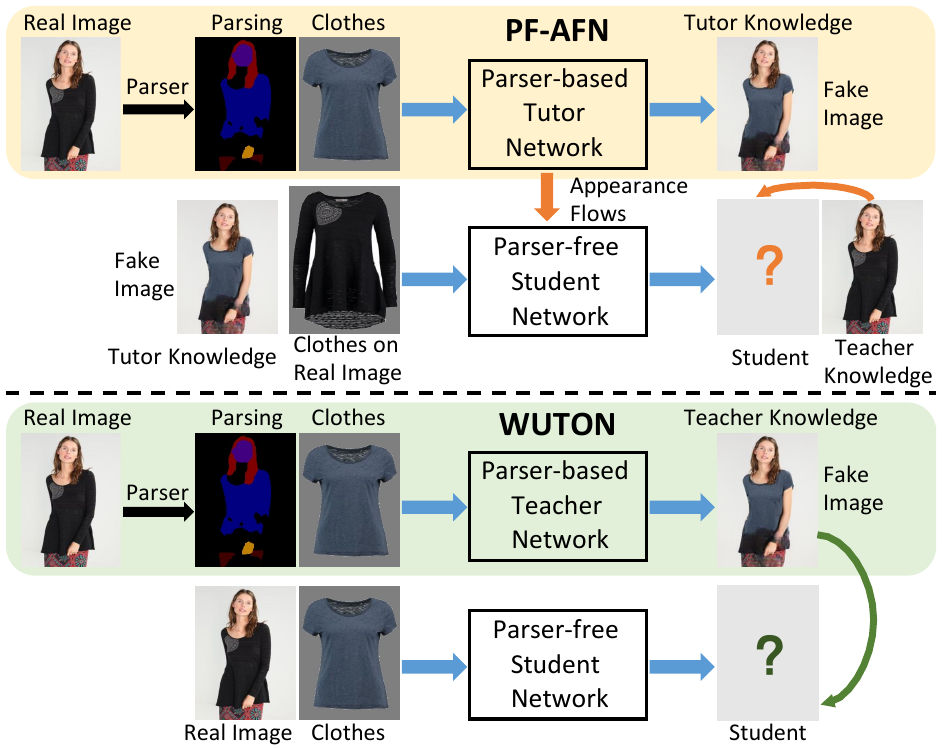}
		\end{center}
		\vspace{-3mm}
		\caption{The comparison between WUTON~\cite{parser_free} and PF-AFN. WUTON treats a parser-based network as a ``teacher" network and distills the fake person image produced by the teacher to a parser-free ``student" network, making the student directly mimic the try-on ability of the parser-based teacher. In comparison, our PF-AFN treats the fake person image as ``tutor knowledge" and uses it as the input of the parser-free ``student" network, which is supervised by the real person image (teacher knowledge). The parser-based ``tutor" network further distills the appearance flows between the person image and the clothes image, facilitating the high-quality image generation in the ``student" network.} 
		\vspace{-10pt}
		\label{fig:our_wuton}
	\end{figure}

	To reduce the dependency of using accurate masks to guide the try-on models, a recent pioneering work WUTON~\cite{parser_free} presented the first parser-free network without using human segmentation for virtual try-on.
	%
	Unfortunately, \cite{parser_free} has an inevitable weakness in its model design. 
	As shown in the bottom of Fig.\ref{fig:our_wuton}, WUTON employed a conventional knowledge distillation scheme by 
	treating a parser-based model (\ie a try-on network that requires human segmentation) as a ``teacher'' network, and distilling the try-on images (\ie fake person images) produced by the teacher to a parser-free ``student'' network, which does not use segmentation as input. This is to make the parser-free student directly mimic the try-on ability of the parser-based teacher.
	However, the generated images of the parser-based teacher have large artifacts (Fig.\ref{fig:compare}), thus using them as the teacher knowledge to supervise the student model produces unsatisfactory results since the image quality of the student is bounded by the parser-based model.

	To address the above problems, this work proposes a new perspective to produce highly photo-realistic try-on images without human parsing, called Parser Free Appearance Flow Network (PF-AFN), which employs a novel ``teacher-tutor-student'' knowledge distillation scheme.
	%
	As shown at the top of Fig.\ref{fig:our_wuton}, instead of treating the parser-based model as the teacher,  PF-AFN only treats it as a ``tutor'' network that may produce unrealistic results (\ie tutor knowledge), which need to be improved by a real teacher.
	%
	The key is to design where the teacher knowledge comes from.
	To this end, PF-AFN treats the fake person image (tutor knowledge) as input of the parser-free student model, which is supervised by the original real person image (teacher knowledge), making the student mimic the original real images.
	This is similar to self-supervised learning, where the student network is trained by transferring the garment on the real person image to the fake person image produced by the parser-based model. In other words, the student is asked to change the clothes on the fake person image to the clothes on the real person image, enabling it to be self-supervised by the real person image that naturally has no artifacts. In this case, the images generated by our parser-free model significantly outperform its previous counterparts.

	To further improve image quality of the student, other than using real images as supervisions, we formulate knowledge distillation of the try-on problem as distilling the appearance flows between the person image and the garment image, facilitating to find dense  correspondences between them to generate high-quality images.

	Our work has three main \textbf{contributions}.
	First, we propose a ``teacher-tutor-student'' knowledge distillation scheme for the try-on problem, to produce highly photo-realistic results without using human segmentation as model input, completely removing human parsing.
	Second, we formulate knowledge distillation in the try-on  problem as distilling appearance flows between the person image and the garment image, which is important to find accurate dense correspondences between pixels to generate high-quality images.
	Third, extensive experiments and evaluations on the popular  datasets demonstrate that our proposed method has large superiority compared to the recent state-of-the-art approaches both qualitatively and quantitatively.
	
	\begin{figure*}[t]
		\begin{center}
			\includegraphics[width=1\linewidth]{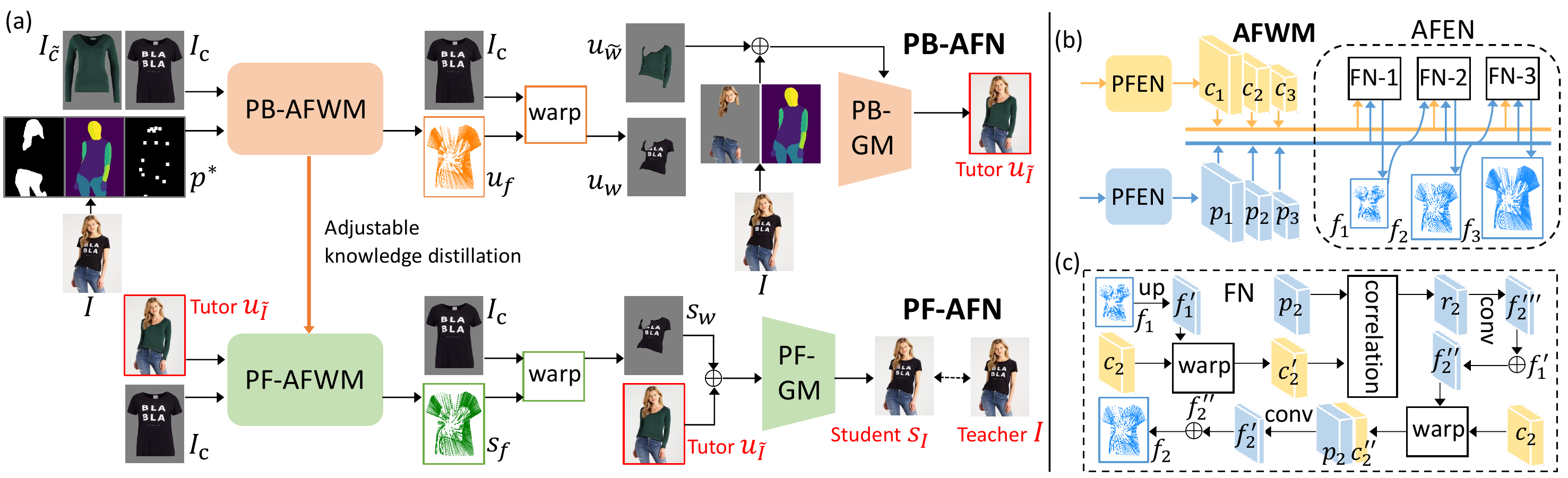}
		\end{center}
		\vspace{-3mm}
		\caption{The training pipeline of PF-AFN. The training data is the clothes image $I_c$ and the image $I$ of a person wearing the clothes. We obtain $p^*$ from the person image $I$ as the parser-based inputs. Given $p^*$, the parser-based network PB-AFN randomly selects a different clothes image $I_{\widetilde{c}}$ to synthesize the fake image ${u}_{\widetilde{I}}$ as the tutor. We use the tutor ${u}_{\widetilde{I}}$ and the clothes image $I_c$ as inputs to train the parser-free network PF-AFN, where the generated student ${s}_I$ is directly supervised by the real image $I$. Furthermore, PB-AFN estimates the appearance flows $u_f$ between $I_c$ and $p^*$, and distills the appearance flows to PF-AFN through the adjustable knowledge distillation. During inference, a target clothes image and a reference person image will be fed into PF-AFN to generate the try-on image, without the need of human parsing results or human pose estimations.}   
		\vspace{-10pt}
		\label{fig:pipeline}
	\end{figure*}
	
	\section{Related Work}
	{\flushleft \bf Virtual Try-on.}
	Existing deep learning based methods on virtual try-on can be classified as 3D model based approaches~\cite{design,animation,animation2019,clothcap} and 2D image based ones~\cite{viton,cpvton,clothflow,vtnfp,cpvton_plus,ACGPN,parser_free}.
	As the former require additional 3D measurements and more computing power, 2D image based approaches are more broadly applicable.
	Since available datasets~\cite{viton,MPV} for 2D image try-on only contain unpaired data (clothes and a person wearing the clothes), previous methods ~\cite{viton,cpvton,clothflow,vtnfp,cpvton_plus,ACGPN} mainly mask the clothing region of the person image and reconstruct the person image with the corresponding clothes image, which require accurate human parsing.
	When parsing results are inaccurate, such parser-based methods generate visually terrible try-on images with noticeable artifacts.
	WUTON~\cite{parser_free} recently proposes a pioneering parser-free approach, but makes the quality of the generated image from a parser-free network bounded by fake images from a parser-based network. 

	{\flushleft \bf Appearance Flow.}
	Appearance flow refers to 2D coordinate vectors indicating which pixels in the source can be used to synthesize the target. It motivates visual tracking~\cite{song2017crest}, image restorations~\cite{Liu2019MEDFE,wang2020rethinking} and face hallucination~\cite{song2019joint}.
	Appearance flow is first introduced by \cite{appearance} to synthesize images of the same object observed from arbitrary viewpoints.
	The flow estimation is limited on the non-rigid clothing regions with large deformation.
	\cite{intrinsic} uses 3D appearance flows to synthesize a person image with a target pose, via fitting a 3D model to compute the appearance flows as supervision, which are not available in 2D try-on.
	
	{\flushleft \bf Knowledge Distillation.}
	Knowledge distillation leverages the intrinsic information of a teacher network to train a student network, which was first introduced in~\cite{distill} for model compression.
	As introduced in~\cite{zhao2020knowledge}, knowledge distillation has also been extended as cross-modality knowledge transfer, where one model trained with superior modalities (\ie multi-modalities) as inputs intermediately supervises another model taking weak modalities (\ie single-modality) as inputs, and the two models can use the same network.

	\section{Proposed Approach}
	We propose an approach to produce highly photo-realistic try-on images without human parsing, called Parser Free Appearance Flow Network (PF-AFN), which employs a novel ``teacher-tutor-student" knowledge distillation scheme.
	We further formulate knowledge distillation of the try-on problem as distilling the appearance flows between the person image and the clothes image.
	We first clarify the overall training scheme with the ``teacher-tutor-student" knowledge distillation in Sec.~\ref{sec:pipeline}. 
	We use an appearance flow warping module (AFWM) to establish accurate dense correspondences between the person image and the clothes image, and a generative module (GM) to synthesize the try-on image, which are introduced in detail in Sec.~\ref{sec:flow} and Sec.~\ref{sec:gm}.
	At last, we describe how we distill the appearance flows to generate high-quality images in Sec.~\ref{sec:kd}.

	\subsection{Network Training} \label{sec:pipeline}
	As shown in Fig.~\ref{fig:pipeline}, our method contains a parser-based network PB-AFN and a parser-free network PF-AFN.
	We first train PB-AFN with data $( I_c, I)$  following the existing methods~\cite{cpvton,clothflow,ACGPN}, where $I_c$ and $I$ indicate the image of the clothes and the image of the person wearing this clothes. 
	We concatenate a mask containing hair, face, and the lower-body clothes region, the human body segmentation result, and the human pose estimation result as the person representations $p^*$ to infer the appearance flows ${u}_f$ between $p^*$ and the clothes image $I_c$.
	Then the appearance flows ${u}_f$ are used to generate the warped clothes ${u}_w$ with $I_c$.
	Concatenating this warped clothes, the preserved regions on the person image and human pose estimation along channels as inputs, we could train a generative module to synthesize the person image with the ground-truth supervision $I$.

	After training PB-AFN, we randomly select a different clothes image $I_{\widetilde{c}}$ and generate the try-on result ${u}_{\widetilde{I}}$, that is the image of person in $I$ changing a clothes.
	Intuitively, the generated fake image ${u}_{\widetilde{I}}$ is regarded as the input to train the student network PF-AFN with the clothes image $I_c$.
	We treat the parser-based network as the ``tutor" network and its generated fake image as ``tutor knowledge" to enable the training of the student network.
	In PF-AFN, a warping module is adopted to predict the appearance flows ${s}_f$ between the tutor ${u}_{\widetilde{I}}$ and the clothes image $I_c$ and warp $I_c$ to ${s}_w$.
	A generative module further synthesizes the student ${s}_I$ with the warped clothes and the tutor.
	We treat the real image $I$ as the ``teacher knowledge" to correct the student ${s}_I$, making the student mimic the original real image.
	Furthermore, the tutor network PB-AFN distills the appearance flows ${u}_f$ to the student network PF-AFN though adjustable knowledge distillation, which will be explained in Sec.~\ref{sec:kd}.

	\subsection{Appearance Flow Warping Module (AFWM).} \label{sec:flow}
	Both PB-AFN and PF-AFN contain the warping module AFWM, to predict the dense correspondences between the clothes image and the person image for warping the clothes.
	As shown in Fig.~\ref{fig:pipeline}, the output of the warping module is the appearance flows ( \eg ${u}_f$ ), which are a set of 2D coordinate vectors.
	Each vector indicates which pixels in the clothes image should be used to fill the given pixel in the person image.
	The warping module consists of dual pyramid feature extraction network (PFEN) and a progressive appearance flow estimation network (AFEN).
	PFEN extracts two-branch pyramid deep feature representations from two inputs.
	%
	Then at each pyramid level, AFEN learns to generate coarse appearance flows, which are refined in the next level. 
	%
	%
	The second-order smooth constraint is also adopted when learning the appearance flows, to further preserve clothes characteristics, \eg logo and stripe.
	The parser-based warping module (PB-AFWM) and the parser-free warping module (PF-AFWM) have the identical architecture except for the difference in the inputs.
	
	\textbf{Pyramid Feature Extraction Network (PFEN)}
	As shown in Fig.~\ref{fig:pipeline}~(b), PFEN contains two feature pyramid networks (FPN)~\cite{fpn} to extract two-branch pyramid features from $N$ levels.
	For the parser-based warping module, the inputs are the clothes image $I_c$ and the person representations $p^*$, while the inputs of the parser-free warping module are the clothes image $I_c$ and the generated fake image ${u}_{\widetilde{I}}$.
	Here we use $\{c_i\}_{i=1}^N$ and $\{p_i\}_{i=1}^N$ to indicate two-branch pyramid features respectively.
	In practice, each FPN contains $N$ stages. 
	It is worth note that we set $N=5$ in our model but show the case $N=3$ in Fig.~\ref{fig:pipeline} for simplicity.

	\textbf{Appearance Flow Estimation Network (AFEN).}
	AFEN consists of $N$ Flow Networks (FN) to estimate the appearance flows from $N$ levels' pyramid features.
	The extracted pyramid features $(c_N,p_N)$ at the highest level $N$ are first fed into FN-1 to estimate the initial appearance flows $f_1$.
	Then $f_1$ and the pyramid features at the $N-1$ level are fed into FN-2 for a finer flow $f_2$.
	The above process continues until the finest flow $f_N$ is obtained, and the target clothes is warped according to $f_N$.

	As illustrated in Fig.~\ref{fig:pipeline}~(c), 
	we carefully design the FN module, which performs pixel-by-pixel matching of features to yield the coarse flow estimation with a subsequent refinement at each pyramid level.
	Take the FN-2 as an example, the inputs are two-branch pyramid features $(c_2,p_2)$, as well as the estimated appearance flow $f_1$ from previous pyramid level.
	The operations in FN can be roughly divided into four stages.
	\textbf{In the first stage}, we upsample $f_1$ to obtain $f^{\prime}_1$, and then $c_{2}$ is warped to $c^{\prime}_{2}$ 
	through sampling the vectors in $c_2$ where the sampling location is specified by $f^{\prime}_1$.
	\textbf{In the second stage}, the correlation maps $r_2$ is calculated based on $c^{\prime}_{2}$ and $p_{2}$.
	In practice, the $j$-th point in $r_2$ is a vector representation, which indicates the result of vector-matrix product between the $j$-th point in $c^{\prime}_{2}$ and the local displacement region centered on the $j$-th point in $p_{2}$.
	In such case, the number of channels of $r_2$ equals to the number of points in the above local displacement region. 
	\textbf{In the third stage}, once $r_2$ is obtained, we then feed it into a ConvNet to predict the residual flow $f^{\prime \prime \prime}_2$, which is added to $f^{\prime}_1$ as the coarse flow estimation $f^{\prime \prime}_2$.
	%
	\textbf{In the fourth stage}, $c_{2}$ is warped to $c^{\prime \prime}_{2}$ according to the newly generated $f^{\prime \prime}_2$.
	Then $c^{\prime \prime}_{2}$ and $p_{2}$ are concatenated and fed into a ConvNet to compute the residual flow $f^{\prime}_2$.
	By adding $f^{\prime}_2$ to $f^{\prime \prime}_2$, we obtain the final flow $f_2$ at pyramid level $2$.

	Intuitively, FN performs matching between two-branch high-level features and a further refinement.
	AFEN progressively refines the estimated appearance flows through cascading $N$ FN, to capture the long-range correspondence between the clothes image and the person image, thus it is able to deal with large misalignment and deformation.

	\begin{figure}[t]
		\begin{center}
			\includegraphics[width=1\linewidth]{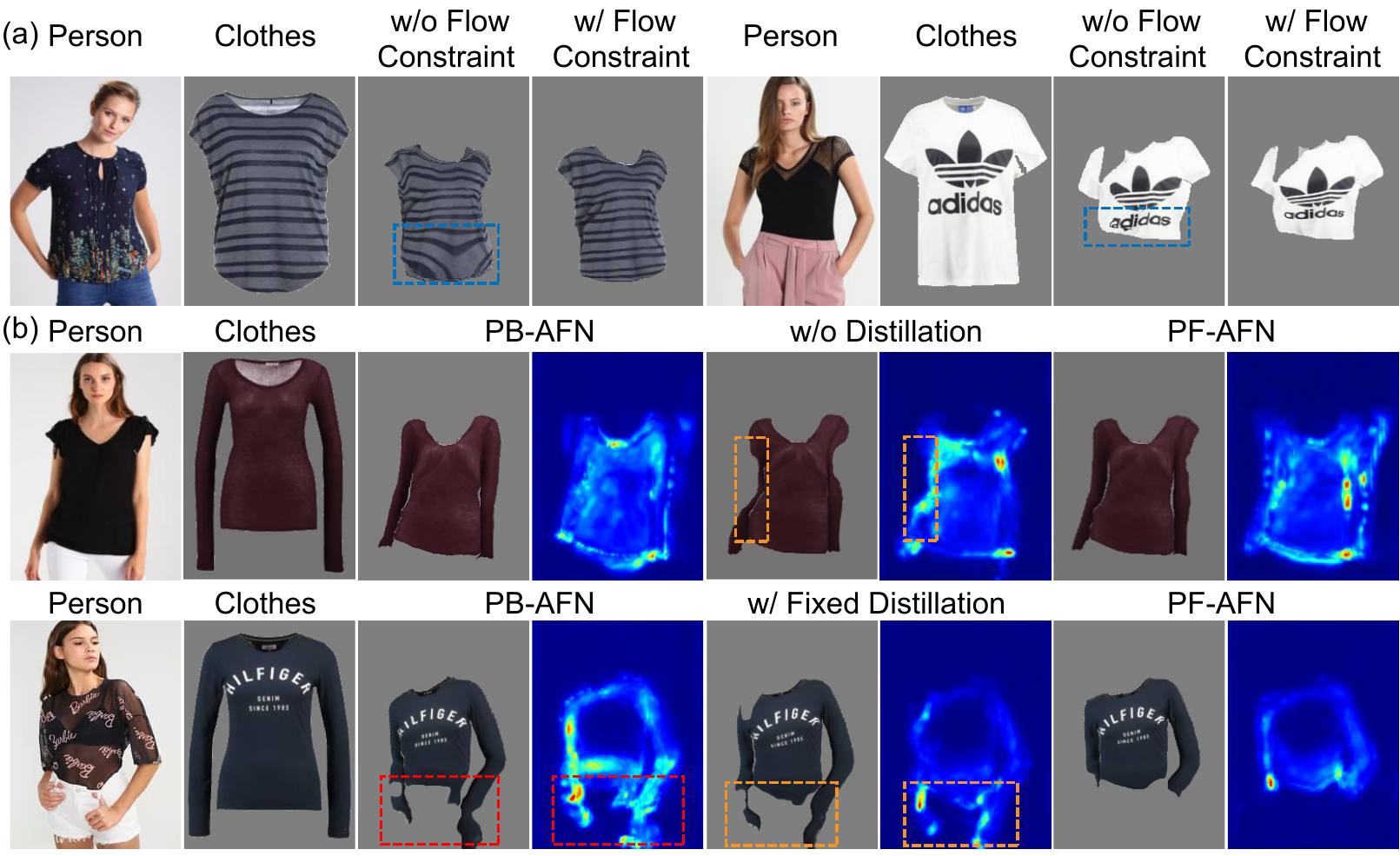}
		\end{center}
		\caption{Comparison between (a) with and without the second-order smooth constraint on appearance flows; (b) when PB-AFN generates accurate warping, our PF-AFN and the model without knowledge distillation; when PB-AFN generates wrong warping, our PF-AFN and the model with fixed knowledge distillation.}
		\vspace{-10pt}
		\label{fig:method_compare}
	\end{figure}

	\textbf{Second-order Smooth Constraint}.
	According to Fig.~\ref{fig:method_compare}, the target clothes usually contain tightly arranged text and the repeating pattern (\eg stripes appear). 
	The appearance flows between the person image and the clothes image need to be predicted accurately, or the minor mistakes should result in very unnatural warping results.
	To better preserve the clothes characteristics, we introduce a second-order smooth constraint to encourage the co-linearity of neighbouring appearance flows.
	The constraint is defined as follows:
	\begin{small}
		\begin{equation}	
			\setlength\abovedisplayskip{4pt}
			\setlength\belowdisplayskip{4pt}
			\mathcal{L}_{sec} = \sum_{i=1}^N\sum_{t}\sum_{\pi\in \mathcal{N}_t}\mathcal{P}( f_i^{t-\pi}  + f_i^{t+\pi} - 2f_i^t )
		\end{equation}
	\end{small}
	where $f_i^t$ denotes the $t$-th point on the flow maps of $i$-th scale (\ie corresponding to the $\{f_i\}_{i=1}^N$ in Fig.~\ref{fig:pipeline}~(b)).
	%
	$ \mathcal{N}_t$ indicates the set of horizontal, vertical, and both diagonal neighborhoods around the $t$-th point. 
	The $\mathcal{P}$ is the generalized charbonnier loss function~\cite{loss}.  
	As illustrated in Fig.~\ref{fig:method_compare} (a), adding $\mathcal{L}_{sec}$ helps maintain the details of the target clothes (\ie the stripes and the characters on the clothes are retained without being distorted).

	\begin{figure*}
		\begin{center}
			\includegraphics[width=1\linewidth]{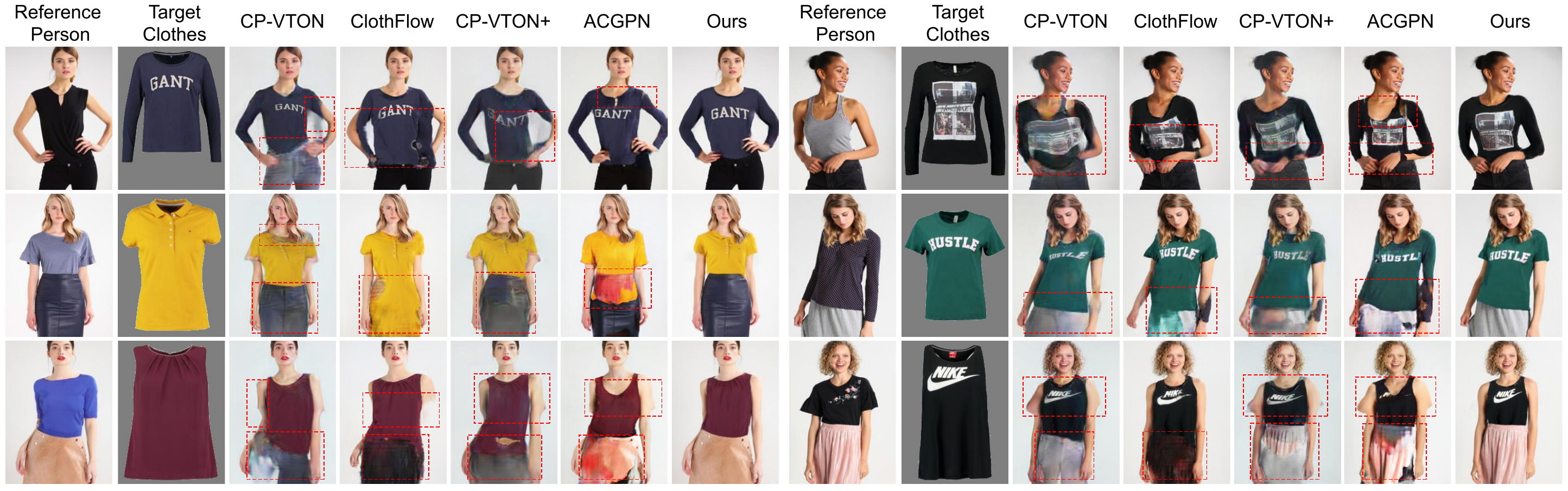}
		\end{center}
		\vspace{-3mm}
		\caption{Visual comparison on VITON dataset. Compared with the recent parser-based methods~\cite{cpvton,clothflow,cpvton_plus,ACGPN}, our model generates more highly-realistic try-on images without relying on human parsing, which simultaneously handles large misalignment between the clothes and the person, preserves the characteristics of both the target clothes and the non-target clothes (\ie skirt), and retains clear body parts.}
		\vspace{-10pt}
		\label{fig:vis_viton}
	\end{figure*}

	\subsection{Generative Module (GM)} \label{sec:gm}
	Both PB-AFN and PF-AFN contain the generative module to synthesize the try-on image.
	The parser-based generative module (PB-GM) concatenates the warped clothes, human pose estimation, and the preserved region on the human body as inputs, while the parser-free generative module (PF-GM) concatenates the warped clothes and the tutor image ${u}_{\widetilde{I}}$ as inputs.
	Both modules adopt the Res-UNet, which is built upon a UNet~\cite{unet} architecture, in combination with residual connections, which can preserve the details of the warped clothes and generate realistic try-on results.

	In the training phase, the parameters of the generative module GM and the warping module AFWM are optimized together by minimizing $\mathcal{L}$, as follows:
	\begin{small}
		\begin{equation}
			\setlength\abovedisplayskip{3pt}
			\setlength\belowdisplayskip{3pt}
			\mathcal{L} = {\lambda}_l\mathcal{L}_l + {\lambda}_p\mathcal{L}_p + {\lambda}_{sec}\mathcal{L}_{sec}
		\end{equation}
	\end{small}
	where $\mathcal{L}_l$ is the pixel-wise L1 loss and $\mathcal{L}_p$ is the perceptual loss~\cite{perceptual} to encourage the visual similarity between the try-on image (\ie the output $s_I$ of the student network) and the real image $I$ as below:
	\begin{small}
		\begin{equation}	
			\setlength\abovedisplayskip{3pt}
			\setlength\belowdisplayskip{-5pt}
			\mathcal{L}_l = {||s_I-I||}_1
		\end{equation}
	\end{small}
	\begin{small}
		\begin{equation}		
			\mathcal{L}_p = \sum_{m}~{||~{\phi}_m(s_I)-{\phi}_m(I)~||}_1
		\end{equation}
	\end{small}
	where ${\phi}_m$ indicates the $m$-th feature map in a VGG-19~\cite{vgg} network pre-trained on ImageNet~\cite{imagenet}.

	\subsection{Adjustable Knowledge Distillation} \label{sec:kd}
	Other than supervising the parser-free student network PF-AFN with the real image $I$, we further distill the appearance flows between the person image and the clothes image, facilitating to find dense correspondences between them.
	As shown in Fig.~\ref{fig:pipeline}~(a), the inputs of the parser-based tutor network PB-AFN include human parsing results, densepose estimations~\cite{densepose} and pose estimations of the input person. 
	In contrast, the input of the student network PF-AFN is only the fake image and the clothes image. 
	Thus, in most cases, the extracted features from PB-AFN usually capture richer semantic information and the estimated appearance flows are more accurate, thus can be used to guide PF-AFN.
	However, as mentioned before, if the parsing results are not accurate, the parser-based PB-AFN would provide totally wrong guidance, making its semantic information and predicted flows irresponsible.
	To address the above issue, we introduce a novel adjustable distillation loss to ensure only accurate representations and predictions are maintained. The definition is as follows:
	\begin{small}
		\begin{equation}
			\setlength\abovedisplayskip{3pt}
			\setlength\belowdisplayskip{3pt}
			\mathcal{L}_{hint} = {\psi}\sum_{i=1}^N{||{u}_{p_i}-{s}_{p_i}||}_2
		\end{equation}
	\end{small}
	\begin{small}
		\begin{equation}	
			\mathcal{L}_{pred} = {\psi}\sum_{i=1}^N{||\sqrt{({u}_{f_i}-{s}_{f_i})^2}||}_1
		\end{equation}
	\end{small}
	\begin{small}
		\begin{equation}	
			{\psi}=
			\begin{cases}
				1, \text{if} \ {||{u}_I-I||}_1<{||{s}_I-I||}_1\\
				0,\text{otherwise}
			\end{cases}
		\end{equation}
	\end{small}
	\begin{small}
		\begin{equation}
			\mathcal{L}_{kd} = {\lambda}_{hint}\mathcal{L}_{hint} + {\lambda}_{pred}\mathcal{L}_{pred}
		\end{equation}
	\end{small}
	where ${u}_{I}$ and ${s}_{I}$ are the generated try-on image from PB-AFN and PF-AFN respectively, $I$ is the real person image.
	$u_{p_i}$ and $s_{p_i}$ are features extracted from the person representation $p^*$ and the fake image ${u}_{\widetilde{I}}$ at the $i$-th scale (\ie corresponding to the $\{p_i\}_{i=1}^N$ in Fig.~\ref{fig:pipeline}~(b)).
	$u_{f_i}$ and $s_{f_i}$ are predicted appearance flows from PB-AFN and PF-AFN at the $i$-th scale (\ie corresponding to the $\{f_i\}_{i=1}^N$ in Fig.~\ref{fig:pipeline}~(b)).
	Specifically, ${\psi}$ is the adjustable factor to decide whether the distillation is enabled by utilizing the teacher to assess the quality of the generated image.
	If the quality of the generated image $u_I$ from the parser-based tutor network does not exceed that of $s_I$ from the parser-free student network (\ie the L1 loss between $u_I$ and $I$ is larger than that between $s_I$ and $I$), the distillation will be disabled. 

	We compare the warped clothes in Fig.~\ref{fig:method_compare} (b) and visualize the activations using the guided prorogation algorithm~\cite{guided}.
	When PB-AFN achieves pleasant performance as shown in the first row, the model without distillation fail to generate accurate warping for the sleeve when it is not activated by the arm.
	When PB-AFN performs poorly as shown in the second row, the model with the fixed distillation (not adjustable distillation) inherits the defects of PB-AFN with erroneous warping to lower-body region when it is activated by the lower-body.
	In both cases, PF-AFN warps the target clothes accurately, which demonstrates the efficiency of the adjustable knowledge distillation. 
	
	\section{Experiments}
	\subsection{Datasets}
	We conduct experiments on VITON~\cite{viton}, VITON-HD~\cite{viton} and MPV~\cite{MPV}, respectively.
	VITON contains a training set of $14,221$ image pairs and a testing set of $2,032$ image pairs, each of which has a front-view woman photo and a top clothing image with the resolution $256 \times 192$.
	Most of previous work in virtual try-on apply this dataset for training and validation.
	VITON-HD is the same as VITON, except that the image resolution is $512 \times 384$.
	It hasn't been tackled before, since it is critically challenging to generate photo-realistic try-on results by giving inputs with high resolutions.
	As a recent constructed virtual try-on dataset, MPV contains $35,687~/~13,524$ person / clothes images at $256 \times 192 $ resolution and a test set of $4175$ image pairs are split out.
	Since there are multiple images of a person wearing the target clothes from different views in MPV, following~\cite{parser_free}, we remove images tagged as back ones since the target clothes is only from the front.
	WUTON~\cite{parser_free} is the only work that conducts experiments on this dataset.

	\begin{figure*}
		\begin{center}
			\includegraphics[width=1\linewidth]{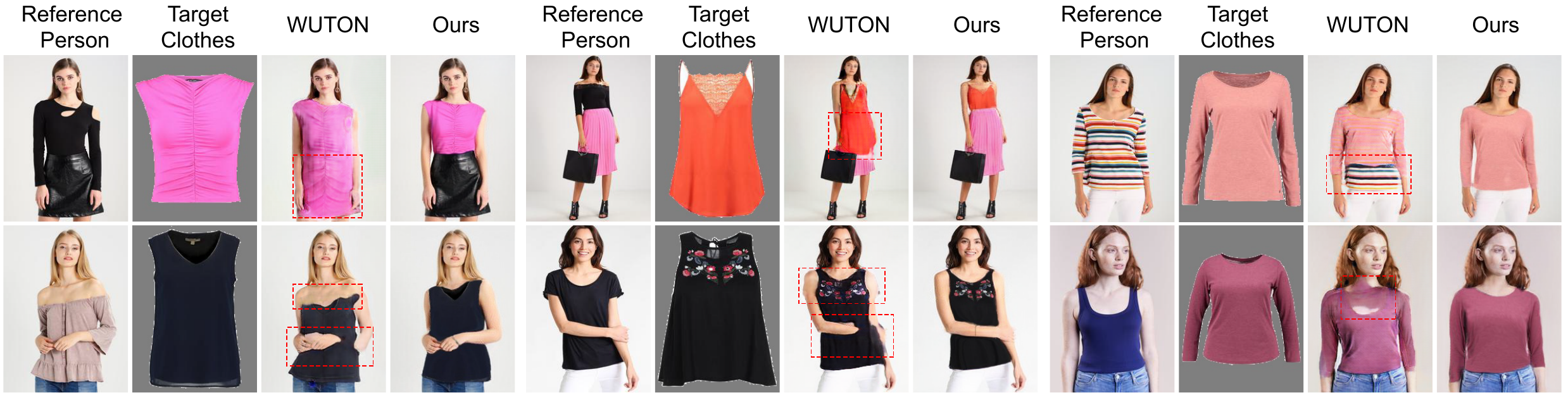}
		\end{center}
		\vspace{-3mm}
		\caption{Visual comparison on MPV dataset with parser-free inputs. Compared with WUTON~\cite{parser_free}, our model generates far more satisfactory results, which warps the target clothes to the person accurately even when the person strikes a complex posture (\ie occlusions and cross-arms), and preserves the characteristics of both the target clothes and the non-target clothes (\ie skirt).}
		\vspace{-10pt}
		\label{fig:vis_mpv}
	\end{figure*}

	\subsection{Implementation Details}
	
	\textbf{Architecture.}
	Both PB-AFN and PF-AFN consist of the warping module (AFWM) and the generative module (GM), where the warping module includes dual pyramid feature extraction network (PFEN) and an appearance flow estimation network (AFEN).
	PFEN adopts the FPN~\cite{fpn} with five layers in practice, and each layer is composed of a convolution with a stride of $2$, followed by two residual blocks~\cite{resnet}.
	AFEN comprises five flow network (FN) blocks, and each FN contains two ConvNets with four convolution layers.
	The generative module has the same structure of Res-UNet~\cite{resunet} in an encoder-decoder style.
	
	\textbf{Training.}
	The training process on three datasets are same.
	we first train PB-AFN with the clothes image and the image of the person wearing the clothes. The parsing results and human pose estimations~\cite{densepose} are also applied in this phase.
	PB-AFN is optimized for 200 epochs with the initial learning rate $3\times 10^{-5}$ and we have ${\lambda}_l =1.0 $, ${\lambda}_p=0.2$, and ${\lambda}_{sec}=6.0$.
	PF-AFN adopts the same training schedule as PB-AFN and uses the same hyper-parameters setting, where ${\lambda}_{hint} =0.04 $ and ${\lambda}_{pred} =1.0 $.

	\textbf{Testing.}
	During test, the reference person image and the target clothes image are given as the input of PF-AFN to generate the image. Additional inputs such as human parsing results and human pose estimations are removed.
	
	\subsection{Qualitative Results}
	\textbf{Results of VITON.} 
	We mainly perform visual comparison of our method with recent proposed parser-based methods in Fig.~\ref{fig:vis_viton}, including CP-VTON~\cite{cpvton}, ClothFlow~\cite{clothflow}, CP-VTON+~\cite{cpvton_plus}, and ACGPN~\cite{ACGPN} .
	As shown in the first row of Fig.~\ref{fig:vis_viton}, 
	when the reference person strikes a complex posture like standing with arms akimbo or two hands blocking in front of the body, large misalignment occurs between the target clothes and the person.
	In such case, baseline models all fail to warp the long-sleeve shirt to the corresponding body region, leading to broken sleeves, sleeves not attached to the arms and distorted embroideries.
	Actually, these methods cannot model the highly non-rigid deformation due to the deficiency of the warping methods, \ie limited degrees of freedom in TPS~\cite{tps}.

	In the second and the third rows of Fig.~\ref{fig:vis_viton}, images generated by baseline methods exist the clear artifacts, such as messy lower-body clothes and top clothes being warped to lower-body region.
	These parser-based models are delicate to segmentation errors because they heavily rely on parsing results to drive the image generation.
	Furthermore, when there exists huge discrepancy between the target clothes and the original clothes on the person ( $\eg$ the person wears a low-collar blouse while the target clothes is high-necked), CP-VTON~\cite{cpvton} and ACGPN~\cite{ACGPN} fail to preserve the characteristics of the target clothes, since they excessively focus on the silhouette of the original clothes during training.
	Moreover, these baseline models are also weak in generating non-target body parts, where obviously fake arms, blurring hands and finger gaps appear on the generated images.

	In comparison, the proposed PF-AFN generates highly-realistic try-on results, which simultaneously handles large misalignment between the clothes and the person, preserves the characteristics of both the target clothes and the non-target clothes, and retains clear body parts.
	Besides the above advantages, benefited from the second-order smooth constraint on the appearance flows, PF-AFN is able to model long-range correspondences between the clothes and the person, avoiding the distortion in logo and embroideries.
	Since we do not mask any information such as clothes or body parts for the input person image during training, PF-AFN can adaptively preserve or generate the body parts, such that the body details can be retained.
	
	\begin{table}
		\begin{center}
			\scalebox{0.85}{
				\begin{tabular}{l|ccc}
					\toprule[2pt]
					\multirow{1}*{Method}&Dataset&FID&Human\\
					\toprule[1.5pt]
					CP-VTON\cite{cpvton}&VITON&24.43&11.15\% / 88.85\%\\
					ClothFlow\cite{clothflow}&VITON&14.43&22.38\% / 77.62\%\\
					CP-VTON+\cite{cpvton_plus}&VITON&21.08&12.62\% / 87.38\%\\
					ACGPN\cite{ACGPN}&VITON&15.67&16.54\% / 83.46\%\\
					PF-AFN (ours) &VITON&\textbf{10.09}& reference\\
					\toprule[1.5pt]
					
					WUTON\cite{parser_free}&MPV&7.927&28.38\% / 71.62\%\\
					PF-AFN (ours)&MPV&\textbf{6.429}& reference\\
					\bottomrule[2pt]
			\end{tabular}}
			\vspace{3pt}
			\caption{Quantitative evaluation results FID~\cite{fid} and user study results. For FID, the lower is the better. For Human result \textbf{``a / b"}, a is the percentage where the compared method is considered better over our PF-AFN, and \textbf{b is the percentage where our PF-AFN is considered better} over the compared method.}
			\vspace{-20pt}
			\label{tab:fid}
		\end{center}
	\end{table}

	\textbf{Results of VITON-HD} 
	The results on VITON-HD are provided in the supplement material.

	\textbf{Results of MPV.} 
	The comparison with WUTON~\cite{parser_free}, which is a pioneer parser-free method, on MPV are shown in Fig.~\ref{fig:vis_mpv}.
	WUTON produces visually unpleasant results with clear artifacts.
	For example, it cannot distinguish the boundary between the top and bottom clothes, making the target top clothes be warped to low-body region.
	In addition, when complicated poses appear in the person images such as occlusions and cross-arms, WUTON generates unnatural results with erroneous warping.
	Since WUTON is supervised by the fake images from a parser-based model that can be misleading, it inevitably achieves unsatisfying performance. 
	In comparison, our PF-AFN can warp the clothes to the target person accurately even in the case of complicated poses and generate high-quality images, which preserves the characteristics of both the target clothes and the non-target clothes (\ie skirt).
	PF-AFN benefits from being supervised by real images as well as finding accurate dense correspondences between the clothes and the person through distilling the appearance flows.
	
	\begin{figure*}
		\begin{center}
			\includegraphics[width=1\linewidth]{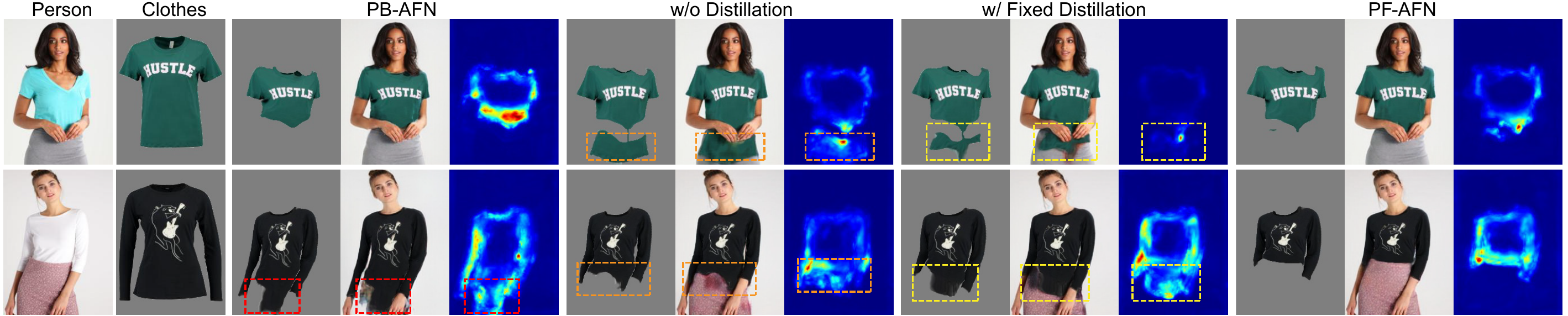}
		\end{center}
		\vspace{-3mm}
		\caption{Ablation studies on the effects of the ``adjustable knowledge distillation''. Given a reference person image and a target clothes image, we show the warped clothes, the try-on image and the visualization of activations using~\cite{guided} for each model.}
		\label{fig:ablation}
	\end{figure*}
	
	\subsection{Quantitative Results}
	For virtual try-on, a target clothes and a reference person image are given to generate the try-on results during the test.
	Since we do not have the ground-truth images (\ie reference person wearing the target clothes), 
	we adopt the Fr\'{e}chet Inception Distance (FID)~\cite{fid} as the evaluation metric following~\cite{parser_free}, which captures the similarity of generated images to real images (\ie reference person images).
	Lower score of FID indicates higher quality of the results.
	We do not use the Inception Score (IS)~\cite{IS} since Rosca \textit{et.al}~\cite{notIS} have pointed out that applying the IS to the models trained on datasets other than ImageNet will give misleading results.

	Table.~\ref{tab:fid} lists the FID scores of CP-VTON~\cite{cpvton}, ClothFlow~\cite{clothflow}, CP-VTON+~\cite{cpvton_plus}, ACGPN~\cite{ACGPN}, WUTON~\cite{parser_free} and proposed PF-AFN on the VITON and MPV dataset.
	Compared with parser-based methods on VITON dataset, our PF-AFN outperforms them by a large margin, showing its great advantage in generating high-quality try-on images without being interfered by parsing results.
	PF-AFN also surpasses WUTON with parser-free inputs on MPV dataset, which demonstrates the superiority of our 'teacher-tutor-student' knowledge distillation scheme.

	\subsection{User Study}
	Although FID can be used as an indicator of the image synthesis quality, it cannot reflect whether the the target clothes are naturally warped with details preserved or the body of the person are retained, so we further conduct a user study by recruiting 50 volunteers in an A / B manner.
	Specially, $300$ pairs from the VITON test set are randomly selected, and CP-VTON~\cite{cpvton}, ClothFlow~\cite{clothflow}, CP-VTON+~\cite{cpvton_plus}, ACGPN~\cite{ACGPN}, PF-AFN each generates 300 images.
	$300$ pairs from the MPV test set are also randomly selected, and WUTON~\cite{parser_free}, PF-AFN each generates $300$ images.
	For each compared method, we have $300$ image groups, where each group contains four images, \ie a target clothes,  a reference person image, two try-on images from the compared method and our PF-AFN, respectively.
	Each volunteer is asked to choose the one with better visual quality.
	%
	As shown in Table.~\ref{tab:fid}, our PF-AFN is always rated better than the other methods with much higher percentage.
	In the A/B test conducted between WUTON and PF-AFN, 71.62\% of the images generated by PF-AFN were chosen by the volunteers to have a better quality.

	\begin{table}[t]
		\vspace{-1mm}
		\begin{center}
			\vspace{-2.5mm}
			\scalebox{0.85}{
				\begin{tabular}{l|c}
					\toprule[1.5pt]
					Different configurations of the AFEN&FID\\
					\toprule[1.5pt]
					An encoder-decoder architecture following~\cite{flownet}.&12.95\\
					\hline
					FN w/o a refinement from ${f_2}^{\prime \prime}$ to $f_2$, only predicting ${f_2}^{\prime \prime}$.&10.79\\
					\hline
					FN w/o ``correlation'' and a refinement from ${f_2}^{\prime \prime}$ to $f_2$,&\multirow{2}*{11.38}\\ 
					only predicting ${f_2}^{\prime \prime}$ by forwarding $p_2$ and $c_2^{\prime}$ to ``conv''.& \\
					\hline
					A single FN w/o cascading $N$ FN modules.&11.90\\
					\toprule[1.5pt]
					PF-AFN with cascading $N$ FN modules. &\bf 10.09\\
					\toprule[1.5pt]
			\end{tabular}}
			\vspace{0pt}
			\caption{Ablation studies of the appearance flow estimation network (AFEN), which consists of flow networks (FN), on VITON. Lower FID indicates better results.}
			\label{tab:afen}
			\vspace{-20pt}
		\end{center}
	\end{table}
	
	\subsection{Ablation Study} 
	\textbf{Adjustable Knowledge Distillation.} We show the ablation studies on the effects of the ``adjustable knowledge distillation''.
	(1) As shown in Fig.~\ref{fig:ablation}, when PB-AFN generates comparatively accurate warping in the first row, the model without knowledge distillation is activated by the low-body and mistakenly warps the top clothes to the low-body region since it does not receive parsing guidance during training. 
	(2) In the second row, when PB-AFN generates erroneous warping caused by the parsing errors, 
	the model with fixed distillation (not adjustable distillation) also generates the failure case because it receives misleading guidance from PB-AFN during training. 
	(3) In contrast, our PF-AFN could generate satisfactory results in both cases. 
	(4) FID on the results predicted by the student network without distillation is 11.40, with fixed distillation is 10.86, and with adjustable distillation is 10.09. Since lower FID indicates better results, the effectiveness of the adjusted knowledge distillation scheme is verified, where only accurate feature representations and predicted flows from a parser-based network will guide the parser-free student network during training.
	
	\textbf{Appearance Flow Estimation Network (AFEN).} We show the ablation studies of the AFEN, which consists of Flow Networks (FN), in Table~\ref{tab:afen}. (1) We use a simple encoder-decoder following~\cite{flownet}. The results are unsatisfying, which indicates that this architecture does not produce accurate appearance flows for clothes warping. (2) We remove refinement, correlation, and cascaded modules of FN, respectively, and get worse results.
	(3) With all of the components, PF-AFN achieves the best performance, which demonstrates the effectiveness of our AFEN.
	
	
	\section{Conclusion}
	In this work, we propose a novel approach, ``teacher-tutor-student" knowledge distillation, to generate highly photo-realistic try-on images without human parsing.
	Our approach treats the fake images produced by the parser-based network (tutor knowledge) as input of the parser-free student network, which is supervised by the original real person image (teacher knowledge) in a self-supervised way.
	Besides using real images as supervisions, we further distill the appearance flows between the person image and the clothing image, to find accurate dense correspondence between them to for high-quality image generation.
	Extensive evaluations clearly show the great superiority of our approach over the state-of-the-art methods.

	\small\noindent\textbf{Acknowledgment} This work is supported by CCF-Tencent Open Fund.
	{\small
		\bibliographystyle{ieee_fullname}
		\bibliography{egbib}
	}
	
\end{document}